\documentclass[a4paper, 10pt, conference]{ieeeconf}      

\IEEEoverridecommandlockouts                              
                                                          
\usepackage{cite}
\usepackage{amsmath,amssymb,amsfonts}
\usepackage{algorithmic}
\usepackage{tikz}
\usepackage{array}
\usepackage{arydshln}
\usepackage{graphicx}
\usepackage{textcomp}
\usepackage{xcolor}
\usepackage{subcaption}

\usepackage[]{caption}
\usepackage[captionskip=2pt]{floatrow}
\newlength\Myfigwd

\usepackage{threeparttable}
\usepackage{multirow}

\usepackage{hyperref}
\def\BibTeX{{\rm B\kern-.05em{\sc i\kern-.025em b}\kern-.08em
    T\kern-.1667em\lower.7ex\hbox{E}\kern-.125emX}}

\newcommand{\realfield}{\hbox{I \kern -.4em R}}
\newcommand {\mb}[1]{\mathbf{#1}}
\newcommand {\bs}[1]{\boldsymbol{#1}}

\newcommand{\T}{^{\mathrm{T}}}

\begin{document}

\title{Admittance Control for Adaptive Remote Center of Motion in Robotic Laparoscopic Surgery}

\author{Ehsan Nasiri \and Long Wang%
	\thanks{The authors are with the Department of Mechanical Engineering, Charles V. Schaefer, Jr. School of Engineering and Science, Stevens Institute of Technology, Hoboken, NJ 07030, USA. 
		{\tt\small \{enasiri, lwang4\}@stevens.edu}}%
    \thanks{This research was supported in part by NSF Grant CMMI-2138896.}
}

\maketitle
\begin{abstract}
	In laparoscopic robot-assisted minimally invasive surgery, the kinematic control of the robot is subject to the remote center of motion (RCM) constraint at the port of entry (e.g., trocar) into the patient's body. During surgery, after the instrument is inserted through the trocar, intrinsic physiological movements such as the patient's heartbeat, breathing process, and/or other purposeful body repositioning may deviate the position of the port of entry. This can cause a conflict between the registered RCM and the moved port of entry. \par
	To mitigate this conflict, we seek to utilize the interaction forces at the RCM. We develop a novel framework that integrates admittance control into a redundancy resolution method for the RCM kinematic constraint. Using the force/torque sensory feedback at the base of the instrument driving mechanism (IDM), the proposed framework estimates the forces at RCM, rejects forces applied on other locations along the instrument, and uses them in the admittance controller. In this paper, we report analysis from kinematic simulations to validate the proposed framework. In addition, a hardware platform has been completed, and future work is planned for experimental validation. 
\end{abstract}

\section{Introduction}
Minimally invasive surgery is a novel approach developed for a variety of procedures like laparoscopic surgery. It utilizes surgical instruments and endoscopes to replace a single large incision with one or several smaller ones. The benefits of avoiding large surgical incisions are well-documented, as these procedures are less painful for patients and result in faster recovery and return to normal activities. Significant improvements in short-term quality of life measures are also reported for using MIS laparoscopic surgeries. In addition enhanced visualization with high-resolution images of the surgical anatomy and pathology provides the surgeon with the opportunity for more precise and accurate surgery\cite{garry2006laparoscopic,Velanovich2000,Veldkamp2005}.

Over the past few decades, scientific and clinical research programs have collaborated to advance robot-assisted surgeries \cite{kwoh1988, Sackier, Ho}. The use of surgical robots, in comparison to traditional procedures, has become increasingly popular, as explained, due to advantages such as higher precision and speed, shorter hospitalization periods, and reduced post-operative pain \cite{Simaan, Kawashima, amodeo2009}. Robot-assisted surgeries have been successfully implemented in various procedures, including laparoscopic Colon and Rectum resection \cite{Antoniou2012, Hubens2003}. \par

A primary consideration in robot-assisted MIS is task completion at the instrument tip (e.g., tissue manipulation), subject to the remote center of motion (RCM) constraint located at the entry port. Robotic systems that incorporate the RCM constraint in their control algorithms require extra degrees of freedom (DoFs) to compensate for the limitations imposed by the RCM \cite{Azimian, Sadeghian}. Programmable RCM offers higher flexibility compared to mechanically designed RCM \cite{palJon, zhang2020}. To address the RCM constraint, a task priority method was developed in \cite{Aghakhani}, while a quaternion-based method was used in \cite{Marinho2014APR}. Studies such as \cite{Locke2007} have employed isotropic-based kinematic optimization to select the RCM location. 
A constrained quadratic optimization approach was proposed in \cite{FUNDA}, and an adaptive motion control guide for needle manipulation to control the virtual RCM was developed in \cite{Boctor}.\par

\begin{figure}[!t]
	\centering
	\includegraphics[width=\columnwidth]{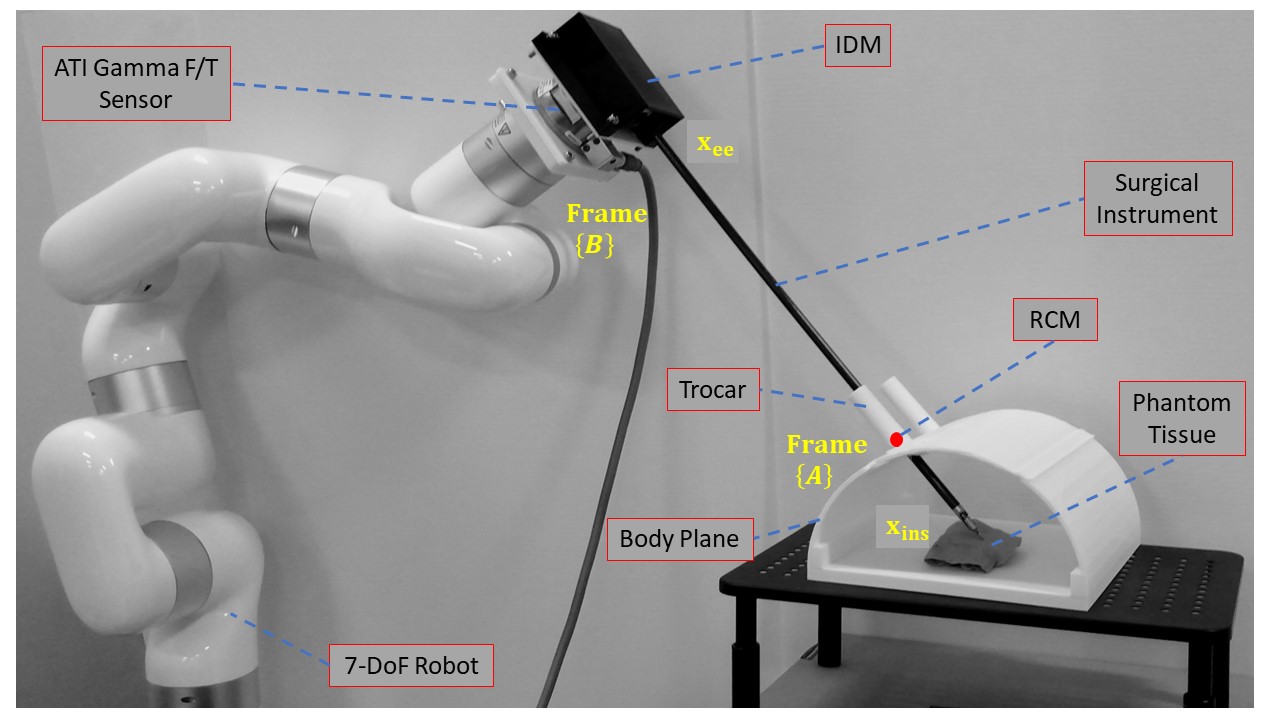}
	\caption{The developed hardware platform includes a 7-DoF robot, a customized IDM, an ATI Gamma F/T sensor, and a surgical instrument.}
	\centering
	\label{hw}
\end{figure}

The majority of papers discussing RCM control treat it as an absolute stationary point (e.g., \cite{Aghakhani, Azimian}), with few exceptions, including \cite{Sadeghian}. Factors such as the patient's breathing and heartbeats may introduce continuous variations in the trocar location during an operation. For instance, movements of the chest area and the abdominal cavity can result in displacements of up to a few centimeters during surgery, as indicated in \cite{Lowanichkiattikul2016, Riviere}. In \cite{Sadeghian}, the proposed methods allowed the RCM to have a velocity, but this required estimating the trocar's velocity using infrared markers or visual sensors. We propose using the interaction force at the RCM location to drive RCM control. In another recent relevant work, \cite{fontanelli2020external}, the authors demonstrated a promising force sensor solution placed at the trocar position. We propose an alternative force/torque sensor placement, offering benefits in terms of convenience for sterilization and capital equipment maintenance. \par

In this paper, we develop a novel framework integrating admittance control with a redundancy resolution method. This approach satisfies the RCM constraint while accommodating changes in the RCM location driven by external forces. This paper introduces several unique contributions:
\begin{itemize}
	\item A novel kinematic framework that integrates admittance control at RCM location with an augmented Jacobian approach.
	\item A robust estimation method for forces at RCM via force/torque sensory feedback at the base of the IDM.
	\item Validations were carried out in simulations on a hardware-developed system that includes a 7-DoF manipulator, a customized IDM, and an off-the-shelf surgical instrument.
\end{itemize}

This paper is outlined as follows. Section~\ref{Kinematic} provides a complete RCM constraint modeling and formulation, followed by a model assumption for the RCM interaction forces during surgery. It then discusses the integration of an admittance controller and a redundancy resolution method for the RCM kinematic constraint. Finally a robust estimation method for forces at RCM through force/torque sensory feedback is presented. In Section~\ref{simulations}, we validate the proposed  control algorithm through simulations in \emph{ROS2} and \emph{MATLAB}.\par

\section{Kinematic Modeling} \label{Kinematic}
\subsection{Nomenclature}
The robotic system includes a robot manipulator that holds an IDM that drives a surgical instrument. In this work, we mainly focus on the commands for the robot manipulator. We use the term \textit{end-effector} (\emph{ee}) for the robot manipulator and the term \emph{instrument} (\emph{ins}) for the instrument wrist center point (shown in Fig.~\ref{hw}and \ref{RCM_schematic}). In other words, an F/T sensor is installed after the end-effector, and the IDM is mounted on the F/T sensor. Throughout the paper, we adopt the nomenclature in Table~\ref{tab:nomenclature}.
\begin{table}[ht]
	\centering
	\caption{Nomenclature for Kinematics}
	\label{tab:nomenclature}
	{\renewcommand{\arraystretch}{1.2}
		\footnotesize
		\begin{tabular}{m{.15\columnwidth} m{.65\columnwidth}}
			\hline
			Symbol & \vspace{1mm} Description \vspace{1mm} \\
			\hline
			Frame~\{A\},  Frame~\{B\}
			& \vspace{1mm} Frames for Remote Center of Motion and  the F/T sensor.\vspace{1mm}\\
			\hdashline
			$\mb{{x}}_{\text{trocar}}$, $\mb{{x}}_{\text{rcm}}$
			& \vspace{1mm} Trocar and RCM positions \\
			\hdashline
			$\mb{{x}}_{\text{ee}}$
			&\vspace{1mm} Robot \emph{end-effector} position\\
			\hdashline
			$\mb{{x}}_{\text{ins}}$
			&\vspace{1mm} Wrist center position of the surgical \emph{instrument}  \\
			\hdashline
			$\mb{d}_\text{ins}$
			&\vspace{1mm}  Vector that points from $\mb{{x}}_{\text{ee}}$ to $\mb{{x}}_{\text{ins}}$ \\
			\hdashline
			$\mb{{J}}_{\text{rcm}}$
			&\vspace{1mm}  Jacobian for the RCM  \\
			\hdashline
			$\mb{{J}}_{\text{ee}}$
			&\vspace{1mm}  Jacobian for the end-effector  \\
			\hdashline
			$\mb{{J}}_{\text{ins}}$
			&\vspace{1mm}  Jacobian for the instrument point  \\
			\hdashline
			$\mb{{J}}_{\text{total}}$
			&\vspace{1mm}  Total augmented Jacobian  \\
			\hdashline
			$\mb{{e}}_{\text{tot}}$
			&\vspace{1mm}  Total augmented error  \\
			\hdashline
			$\mb{{K}}_{\text{ins}}$,  $\mb{{K}}_{\text{rcm}}$
			&\vspace{1mm}  Diagonal $3\times3$ gains for the instrument position and for the RCM constraint \\
			\hdashline
			$\mb{{K}}_{\text{env}}$,  $\mb{{B}}_{\text{env}}$
			&\vspace{1mm}   Diagonal $3\times3$ stiffness and damping for environment compliance \\
			\hdashline
			$\mb{K}_{\text{adm}}$, $\mb{K}_{\text{adm}\perp}$
			&\vspace{1mm}   An scalar admittance gain and it's projection  \\  
			\hdashline
			$\mb{{f}}_{\text{rcm}}$
			&\vspace{1mm}  Actual force exerted from patient's body at RCM point \\
			\hdashline
			$\mb{{f}}_b$, $\mb{{m}}_b$
			&\vspace{1mm}  Force and moment at the IDM base \\
			\hdashline
			$\bar{\mb{f}}_b$, $\bar{\mb{m}}_b$
			&\vspace{1mm}  Measured force/torque at F/T sensor frame \\
			\hdashline
			$\hat{\mb{{f}}}_{\text{rcm}}$
			&\vspace{1mm}  Estimated Force exerted from patient's body at RCM point by F/T sensor  \\
			\hdashline
			$\mb{{f}}_{\text{e,rcm}}$
			&\vspace{1mm}  RCM force tracking error\\
			\hdashline
			$\mb{{f}}_{\text{ins}}$
			&\vspace{1mm}  Force acting upon the instrument distal end (not around RCM)\\
			\hdashline
			$\bs{\Omega}$
			&\vspace{1mm}   The constructed projection matrix  \\
			\hdashline
			$\dot{\mb{{x}}}_{\text{cmd}}$
			&\vspace{1mm}  Augmented command velocity vector \\
			\hdashline
			$\mb{G}$
			&\vspace{1mm}   A diagonal $3\times3$ gain for the command vector, $\dot{\mb{{x}}}_{\text{cmd,adm}}$   \\
			\hdashline
			${l}$
			&\vspace{1mm}   The surgical instrument Length  \\
			\hdashline
			$\eta$
			&\vspace{1mm}   The interpolation variable of RCM point along the instrument shaft, $\eta\in(0,1)$.  \\  
			\hline
		\end{tabular}
	}
\end{table}

\subsection{Kinematic Constraints: RCM \& Instrument}\label{RCM formulation}
In this work, we build our framework upon previous studies, such as \cite{Aghakhani, ROY}, to enforce both the RCM constraint and the instrument's commanded motion constraint. \par
\begin{figure}[!t]
	\centering
 	\includegraphics[width=\columnwidth]{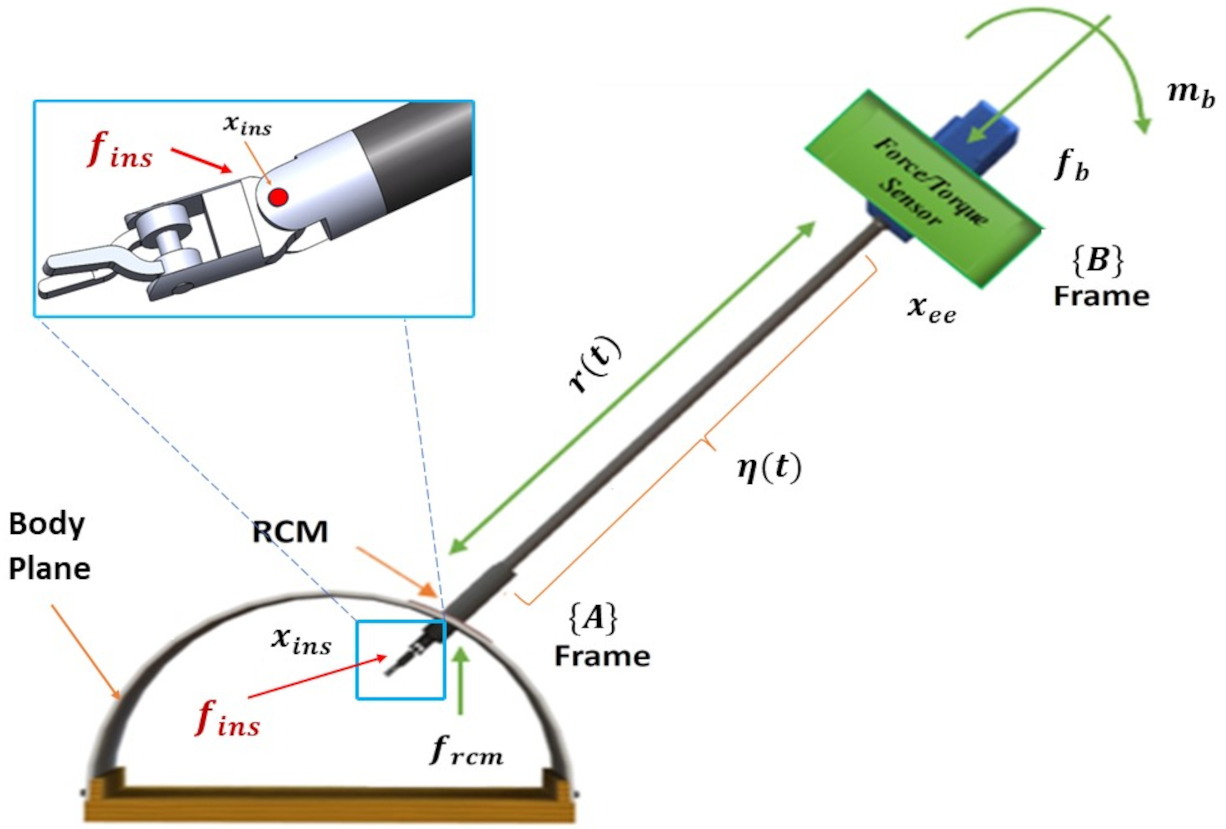}
	\caption{Schematic representation of RCM on a surgical robot's end-effector subject to external forces.}
	\label{RCM_schematic}
\end{figure}

The position of the RCM point is defined using an interpolation variable $\eta$: 
\begin{equation} 
	\mb{x}_{\text{rcm}} = \mb{x}_\text{ee} + \eta \,(\mb{x}_\text{ins} - \mb{x}_\text{ee}), \quad \eta\in(0,1)
	\label{eq:def_RCM}
\end{equation}
where $\mb{x}_\text{ee}$ denote the \emph{end-effector} position while $\mb{x}_\text{ins}$ the \emph{instrument} position. Thereby, \eqref{eq:def_RCM} conveniently defines the RCM point in between the end-effector and the instrument, shown in Fig.~\ref{RCM_schematic}. It is worth noting that the RCM location specified by $\eta$ does not necessary agree with the trocar position.\par

Taking the derivatives of both sides of \eqref{eq:def_RCM}, we arrive at:
\begin{equation}
	\mb{\dot{x}}_{\text{rcm}} =
	\underbrace
	{\;
		\Big[
		\begin{array}{c;{2pt/2pt}c}
			(\mb{J}_\text{ee} + \eta \,(\mb{J}_\text{ins} - \mb{J}_\text{ee}) &  \mb{d}_\text{ins}
		\end{array} \Big]\;}_{\triangleq\; \mb{J}_{\text{rcm}}({q},\eta)}
	\; 
	\begin{bmatrix} \mb{\dot{q}} \\ \dot{\eta} \end{bmatrix}
	\label{RCM_velocity}
\end{equation}
where $\mb{d}_\text{ins}$, defined as below, is the relative instrument position w.r.t the end-effector 
\begin{equation}
	\mb{d}_\text{ins} = \mb{x}_\text{ins} - \mb{x}_\text{ee}   
\end{equation}
We therefore can define the RCM Jacobian as: 
\begin{equation}
	\mb{J}_{\text{rcm}}({q},\eta) ={\;
		\Big[
		\begin{array}{c;{2pt/2pt}c}
			(\mb{J}_\text{ee} + \eta \, (\mb{J}_\text{ins} - \mb{J}_\text{ee}) &  \mb{d}_\text{ins}
		\end{array} \Big]\;}\label{RCM_jacob}
\end{equation}
where matrix $\mb{J}_{\text{ins}}$ is the manipulator Jacobian that maps joint velocities to the instrument (denoted by $\text{ins}$) velocity:
\begin{equation}
	\mb{\dot{x}}_{\text{ins}}= \mb{J}_{\text{ins}}(\mb{q}) \;\mb{\dot{q}}
	\label{ins_velocity}
\end{equation}\par
In this work, we specify the RCM constraint and the instrument commanded motion constraint using velocity conditions:
\begin{align}
	s.t.\quad
	& \mb{J}_\text{rcm}(\mb{q},\eta) 
	\begin{bmatrix}
		\dot{\mb{q}}\\
		\dot{\eta}
	\end{bmatrix} \; = \dot{\mb{x}}_\text{rcm,cmd} \label{eqn:RCM_constraint}\\[2pt]
	s.t.\quad & \mb{J}_\text{ins}(\mb{q}) 
	\;\dot{\mb{q}}
	\;= \dot{\mb{x}}_\text{ins,cmd}\label{eqn:ins_constraint}
\end{align}
Combining both constraints using an augmented Jacobian definition
\begin{equation}
	s.t.\quad
	\mb{J}_\text{total}(\mb{q},\eta) 
	\begin{bmatrix}
		\dot{\mb{q}}\\
		\dot{\eta}
	\end{bmatrix} \; = \dot{\mb{x}}_\text{cmd} \label{eqn:total_constraint} 
\end{equation}
Where,
\begin{align}
	& \mb{J}_{\text{{total}}} = \left[\begin{array}{c}
		\mb{J}_{\text{{ins}}} \quad \mb{0}_{3\times1} \\
		\hdashline
		\mb{J}_{\text{{rcm}}}
	\end{array}\right]
	\label{eq:total_jacobian} \\[2pt]
	&     
	\dot{\mb{x}}_{\text{cmd}} = 
	\left[\begin{array}{c}
		\dot{\mb{x}}_{\text{cmd,ins}}\\
		\hdashline
		\dot{\mb{x}}_{\text{cmd,rcm}}
	\end{array}\right]
	\label{eqn:x_cmd_velocity}
\end{align}

Through the next few sections, we will develop a framework that allows an admittance force controller to drive the RCM velocity constraint.

\subsection{Redundancy Resolution}
A general solution that satisfies the constraint in \eqref{eqn:total_constraint} can be obtained as:
\begin{equation}
    \begin{split}
        \begin{bmatrix}
            \dot{\mathbf{q}} \\
            \dot{\eta}
        \end{bmatrix}_{\text{cmd}} = \mathbf{J}_{\text{total}}^{\dagger} \mathbf{\dot{x}}_{\text{cmd}} + \left(\mathbf{I}-\mathbf{J}_{\text{total}}^{\dagger} \mathbf{J}_{\text{total}}\right) \mathbf{w}
    \end{split}
    \label{redundancy_without_admt} 
\end{equation}

It is worth noting that the system should have 2-DoF redundancy, considering that the manipulator has 7-DoF, the instrument command motion constraint is 3-DoF and the RCM constraint is 2-DoF.

The redundancy is then resolved by the choice of a vector $\mb{w}$ in \eqref{redundancy_without_admt}, which represents an arbitrary choice that can satisfy the task constraint. It can be engineered as a gradient of a cost function of interest. We choose a cost function as below:
\begin{equation}
	L= \tfrac{1}{2}({\eta}-\eta_0)\T ({\eta}-\eta_0)\label{weight}
\end{equation}
where \(\eta_0\) is a fixed value (e.g., in this paper 0.25), to ensure that the value of ${\eta}$ stays as close to that value as possible. Then, the null-space cost gradient is:
\begin{align}
	\mb{w}=&\quad \nabla_{(\mb{q},\,{\eta})} L, \quad  
	\text{or} \\
	\mb{w}=&\quad \begin{bmatrix} 
		\begin{array}{c}0, 0, 0, 0, 0, 0, 0, {\eta}-{\eta{_0}} \end{array} \end{bmatrix}\T 
	\label{eq:24}
\end{align}

\subsection{Model Assumptions for RCM Interaction Forces}\label{force modeling assumption}
\begin{figure}[!t]
	\centering
	\includegraphics[width=\columnwidth]{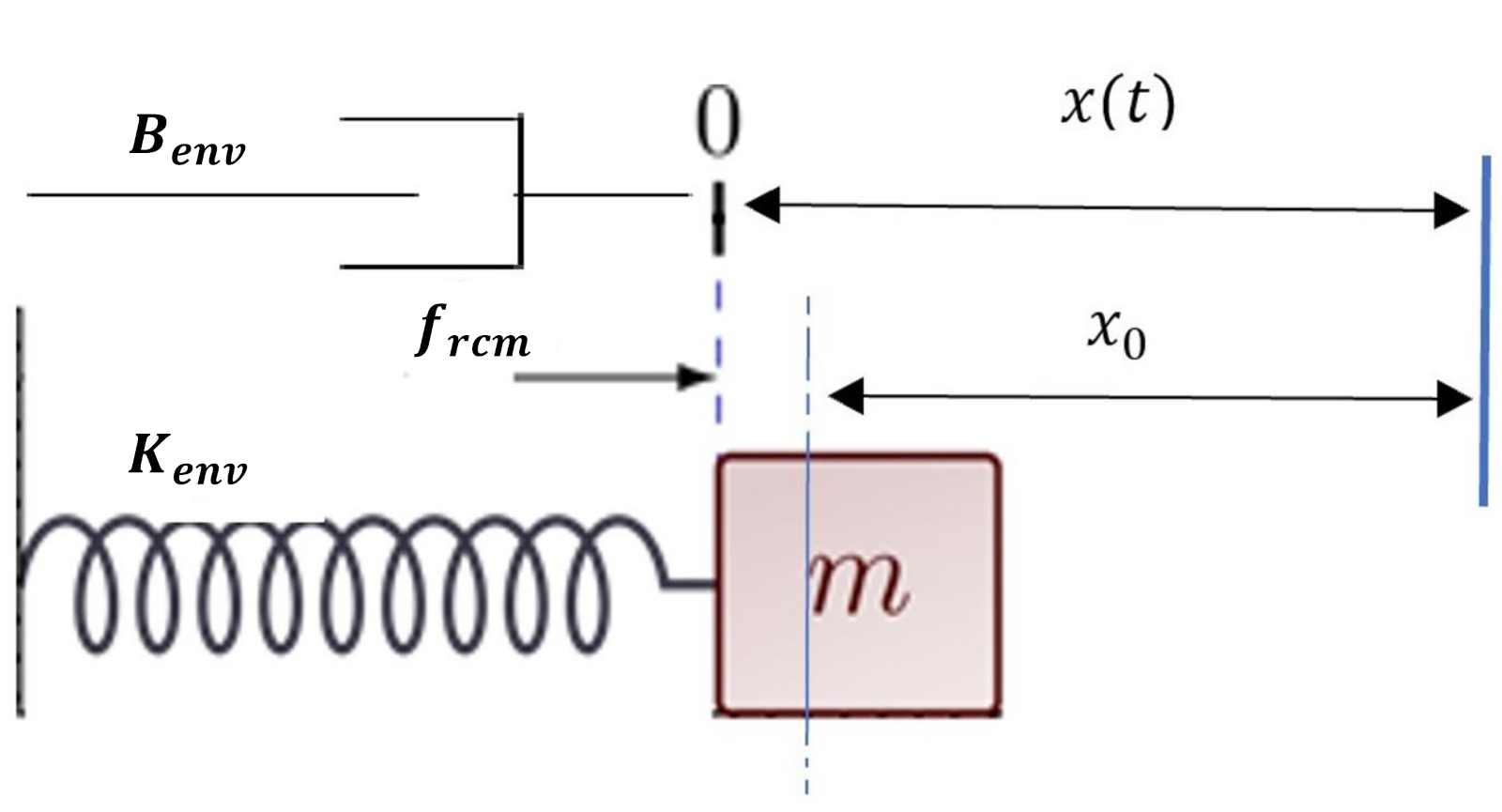}
	\caption{The stiffness-damping behavior of a 1-DoF model for the end-effector at the RCM point during operation.}
	\label{stiffness_damping}
\end{figure}

During a laparoscopic robot-assisted MIS operation, due to potential body motion of a patient, force can be exerted at the RCM location. We assume that, at the RCM location, the instrument shaft is in contact with an environment that has unknown, but linear, and time-invariant compliance, as depicted in Fig.~\ref{RCM_schematic}. In a one-dimensional example, the force-displacement model will be similar to ${f}_{\text{rcm}} = {k}_{\text{env}}(x-x_0)+{b}_{\text{env}}(\dot{x}-\dot{x}_0)$, illustrated in Fig.~\ref{stiffness_damping}. The parameters ${k}_{\text{env}}$ and ${b}_{\text{env}}$ represent the environment stiffness and damping.\par

In a three-dimensional case, we model the interaction forces at the RCM location as a spatial spring damper system. When there is a displacement between the trocar and the instrument shaft, a force can be approximated as below:
\begin{align}
	& \mb{f}_\text{rcm}=\mb{K}_{\text{env}} \; \mb{x}_{\text{e}} + \mb{B}_{\text{env}} \; \mb{\dot{x}}_{\text{e}} \label{eqn:f_rcm_model}
	\\    
	& \mb{x}_{\text{e}} = \mb{x}_{\text{trocar}} - \mb{x}_{\text{rcm}} \label{position error}\\
	& \mb{\dot{x}}_{\text{e}} = \mb{\dot{x}}_{\text{trocar}} - \mb{\dot{x}}_{\text{rcm}} \label{velocity error}
\end{align}

In simulations of this work, we use \eqref{eqn:f_rcm_model} to compute the true RCM interaction forces, $\mb{f}_\text{rcm}$. And the admittance controller uses the estimated RCM interaction forces, $\hat{\mb{f}}_\text{rcm}$, as a sensory feedback.

n\subsection{Integrating Admittance Control}\label{integrate addmittance controller}
We have defined an augmented command velocity constraint $\dot{\mb{x}}_\text{cmd}$ in \eqref{eqn:x_cmd_velocity}, which includes $\dot{\mb{x}}_\text{cmd,rcm}$. In this section, we complete the admittance controller integration by determining the velocity $\dot{\mb{x}}_\text{cmd,rcm}$ using the estimated force $\hat{\mb{f}}_\text{rcm}$.\par
The admittance controller is designed as below:
\begin{align} 
	& \mb{\dot{x}}_{\text{cmd,rcm}} = K_\text{adm} \; \left(\mb{I} - \bs{\Omega}\right) \; \;{\mb{f}}_{\text{e,rcm}}\label{RCM_velocity2}\\[6pt]
	& \bs{\Omega} = {\mb{n}_d}\; \left({\mb{n}_d}\T\right), \qquad \mb{n}_d = \frac{\mb{d}_\text{ins}}{\|\mb{d}_\text{ins}\|} \label{eqn:Omega_proj}\\[2pt]
	& {\mb{f}}_{\text{e,rcm}}=\hat{\mathbf{f}}_{\text{rcm}} - {\mathbf{f}}_{\text{rcm,desired}}\label{force_error}
\end{align}\par 
In \eqref{RCM_velocity2}, an admittance law is proposed. A projection matrix, $\bs{\Omega}$, is constructed as in \eqref{eqn:Omega_proj}, so that the null space projector $\left(\mb{I} - \bs{\Omega}\right)$ can be used to compute a projected force in a plane that is located at RCM and that is perpendicular to the instrument shaft. A scalar admittance gain, $K_\text{adm}$, is used for relating the in-plane force and in-place velocity command. Vector ${\mb{f}}_{\text{e,rcm}}$ is the RCM force tracking error.\par 

For convenience of derivation, we define a projected admittance gain matrix:
\begin{equation}
	\mb{K}_{\text{adm}\perp} = K_\text{adm} \; \left(\mb{I} - \bs{\Omega}\right), \quad \mb{K}_{\text{adm}\perp} \in \realfield^{3\times3}
\end{equation}

We then recall the general solution from before:
\begin{align}
	\begin{split}
		\begin{bmatrix}\mb{\dot{q}} \\ \dot{\eta}\end{bmatrix}_{\text{cmd}} = \mb{J}_{\text {total }}^{\dagger} &  \mb{\dot{x}}_{\text{cmd}} + \left(\mb{I}-\mb{J}_{\text {total}}^{\dagger} \mb{J}_{\text {total }}\right) \mb{w}
	\end{split}
	\tag{\ref{redundancy_without_admt}}
\end{align}
where we substitute the command velocity vector $\mb{\dot{x}}_{\text{cmd}}$ as:
\begin{align}
	& \dot{\mb{x}}_{\text{cmd}}=\begin{bmatrix} \mb{K}_{\text{ins}} & \mathbf{0}_{3 \times 3} \\ \mathbf{0}_{3 \times 3} & \mb{K}_{\text{adm}\perp} \end{bmatrix} \mathbf{e}_{\text{tot}}
	\label{new_command} \\[6pt]
	&
	\mathbf{e}_{\text{tot}} = \left[\begin{array}{c}
		\mathbf{x}_{\text{desired}} - \mathbf{x}_{\text{ins}} \\
		\hdashline
		\hat{\mathbf{f}}_{\text{rcm}} - {\mathbf{f}}_{\text{rcm,desired}}
	\end{array}\right]
	\label{error_new}
\end{align}

In this work $\mb{f}_{\text{rcm,desired}}$ is assumed to be zero. The total error vector, $\mb{e}_\text{tot}$, includes both position errors from the instrument, and the force errors from the RCM location. For convenience, let us define a combined gain matrix $\mb{G}$:
\begin{equation}
	\mb{G}= \begin{bmatrix} \mb{K}_{\text{ins}} & \mathbf{0}_{3 \times 3} \\ \mathbf{0}_{3 \times 3} & \mb{K}_{\text{adm}\perp} \end{bmatrix}
\end{equation}
	
\noindent We rewrite \eqref{redundancy_without_admt} as,
\begin{equation}
	\begin{bmatrix}\mb{\dot{q}} \\ \dot{\eta}\end{bmatrix}_{\text{cmd,adm}} = \mb{J}_{\text {total}}^{\dagger} \;
	\mb{G} \; \mb{{e}}_{\text {tot}}  + 
	\left(\mb{I}- \mb{J}_{\text {total}}^{\dagger} \mb{J}_{\text {total}} \right)  \mb{w}
	\label{eq:16}
\end{equation}
	
This finalized equation, as given in~$\eqref{eq:16}$, is subsequently used to determine the joint space command velocity for the robot manipulator. For instance, it can facilitate tracking the instrument's position while controlling the movement of the RCM point caused by interaction forces.\\

\begin{figure*}[!h]
	\setlength\Myfigwd{0.76\columnwidth}
	\floatbox[{\capbeside\thisfloatsetup{
			capbesideposition={right,center},
			capbesidewidth=\dimexpr\linewidth-\Myfigwd-2em\relax}}]{figure}[\FBwidth]
	{\caption{Simulation plots include: 
			  1) Convergence of the force acting upon the RCM point from the patient’s body with the proposed control algorithm.  
			2) Convergence of the estimated criterion metric, $\hat{\gamma}$, with the applied force loads.
			3) Convergence of position and velocity errors with the exerted force acting upon the RCM point.}
		\label{simulation_plots}}
	{\includegraphics[width=\Myfigwd]{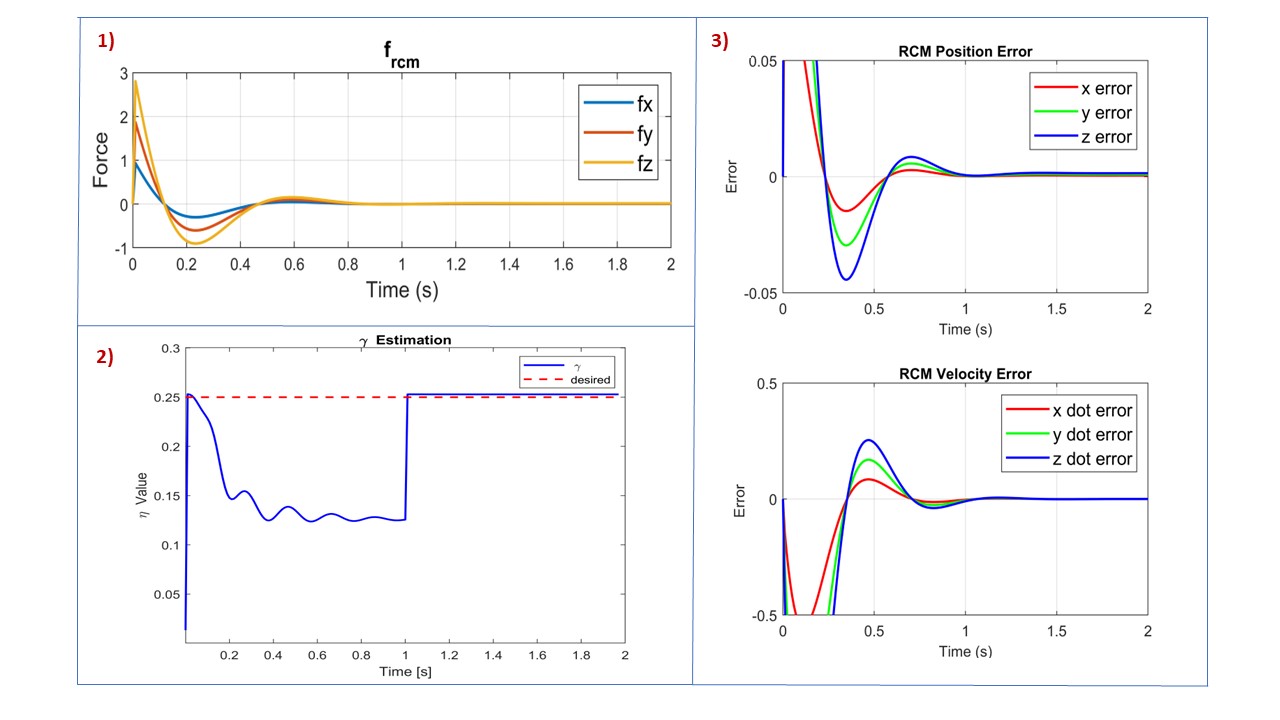}}
	\vspace{-3mm}
\end{figure*}
\subsection{Force Estimation at RCM Location} \label{eta variation formula}
While equipping a force sensor near the instrument shaft and trocar is promising (e.g., \cite{fontanelli2020external}), it poses additional challenges in terms of sterilizing the instrument and/or designing it for disposable use. Alternatively, in this work, we place a force/torque sensor at the base of the IDM. This approach offers convenience for sterilization and capital hardware maintenance. However, it adds complexity as it does not directly measure the interaction forces at the RCM. \par
Thereby, the challenge of our method stems from the possibility of multiple force loading locations along the instrument shaft, causing difficulty in decoupling the forces at different locations using a single force/torque sensory feedback.  We address this problem by simplifying it into two possible cases, referring to Fig.~\ref{RCM_schematic}: 
\begin{itemize}
	\item[1)] A force, $\mb{f}_\text{rcm}$, is applied at the RCM location, $\mb{r}(t)$.
	\item[2)] A force, $\mb{f}_\text{ins}$, is applied at the Instrument location $\mb{d}_\text{ins}(t)$, while $\mb{f}_\text{rcm}$ is applied at $\mb{r}(t)$.
\end{itemize}
While we can extend to define more cases, these two provide reasonable coverage for clinical usage.\\

\setcounter{subsubsection}{-1}
\subsubsection{Criterion to Determine which Case} We introduce a metric, $\hat{\gamma}(t)$, to determine which case to use. To explain the metric, consider the equilibrium of a scenario where there is a force, $\mb{f}_\text{unknown}$ applied on an unknown location along the instrument shaft, $\mb{r}'(t)$.
\begin{align}
	\mb{f}_b + {\mb{f}}_{\text{unknown}}&= \mb{0} \label{eq:18} \\
	\mb{m}_b + \mb{r}'(t) \times {\mb{f}}_{\text{unknown}} &= \mb{0}\\
	\mb{r}'(t) = \gamma(t) \; \mb{d}_\text{ins}(t),   &\phantom{=0}\gamma\in(0,1)
	\label{eqn:criterion_gamma}
\end{align}
Where $\mathbf{f}_b$ and $\mathbf{m}_b$ are the force and moment at the IDM base, as illustrated in Fig.~\ref{RCM_schematic}, and an interpolation variable $\gamma(t)$ is introduced to enforce that the point is along the instrument. We can then use the following estimator for $\gamma(t)$, utilizing the measurements from the F/T sensor:
\begin{equation}
	\hat{\gamma}(t) = \frac{\|\bar{\mb{m}}_b\|}{\|\mb{d}_\text{ins}\times\bar{\mb{f}}_b\|}
 \label{gama_criterion}
\end{equation}
If the estimated interpolation variable, $\hat{\gamma}(t)$, is close to the RCM interpolation variable, $\eta(t)$, then we opt to use \emph{Case 1}.\\
\subsubsection{Single Force Loading by $\mb{f}_\text{rcm}$} In this case, the estimator is simply $\hat{\mb{f}} = - \bar{\mb{f}}_b$. The criterion metric $\hat{\gamma}(t)$ should be estimated and monitored in real-time for each time step.

\subsubsection{Two Forces Loading by $\mb{f}_\text{rcm}$, $\mb{f}_\text{ins}$}
Let us consider the equilibrium that involves both forces from the RCM point and from the instrument point:
\begin{align}
	\mb{f}_b + {\mb{f}}_{\text{rcm}}+ {\mb{f}}_{\text{ins}}&= \mb{0} \label{eqn:force_equilibrium} \\
	\mb{m}_b + \mb{r}(t) \times {\mb{f}}_{\text{rcm}} + \mb{d}_\text{ins}(t) \times {\mb{f}}_{\text{ins}}&= \mb{0} \label{eqn:moment_equilibrium}
\end{align}
where the vector $\mb{r}(t)$ can be expressed using the RCM interpolation variable $\eta(t)$ (shown in Fig.~\ref{RCM_schematic}):
\begin{equation}
	\mb{r}(t) = \eta(t) \; \mb{d}_\text{ins}
	\label{eqn:r_t_definition}
\end{equation}

Let us rewrite the equilibrium in a matrix form using \eqref{eqn:force_equilibrium}, \eqref{eqn:ins_constraint}, and \eqref{eqn:r_t_definition}:
\begin{gather}
	\underbrace{\left[\begin{array}{c;{2pt/2pt}c}
			\mb{I}_{3} & \mb{I}_{3} \vspace{2mm}\\\hdashline
			\phantom{\Bigg[} \Big[\mb{d}_\text{ins}(t)\Big]_\times &
			\eta(t) \Big[\mb{d}_\text{ins}(t)\Big]_\times
		\end{array}
		\right]}_{\triangleq\; \mb{\Gamma}(t)}
	\underbrace{
		\begin{bmatrix}
			\phantom{\Big[}\hspace{-1mm}
			\mb{f}_\text{ins} 
			\hspace{-1mm}\phantom{\Big]}\\
			\phantom{\Big[}\hspace{-1mm}
			\mb{f}_\text{rcm} 
			\hspace{-1mm}\phantom{\Big]}
		\end{bmatrix}
		\raisebox{-7.5mm}{}}_{\triangleq\;\mb{f}_\text{case2}}
	=
	\underbrace{\begin{bmatrix}
			\phantom{\Big[} \hspace{-2mm}
			-\mb{f}_b 
			\hspace{-1mm}\phantom{\Big]}\\
			\phantom{\Big[} \hspace{-2mm}
			-\mb{m}_b 
			\hspace{-1mm}\phantom{\Big]}
		\end{bmatrix}
		\raisebox{-7.5mm}{}}_{\triangleq\;\bs{\xi}_b}
	\\
	\text{Or, }\qquad \bs{\Gamma}(t) \; \mb{f}_\text{case2}\quad = \quad  \bs{\xi}_b
\end{gather}
where $[\mb{v}]_\times$ represents the cross product matrix operator:
\begin{equation}
	[\mb{v}]_\times = 
	\begin{bmatrix}
		0 & -v_3 & v_2 \\
		v_3 & 0 & - v_1 \\
		-v_2 & v_1 & 0
	\end{bmatrix}
\end{equation}

The force estimation matrix $\bs{\Gamma}(t)$ is not full rank, and in a normal case, it will have a rank of 5 because the lower half rows have the cross product matrix operators.  Therefore, the estimation that decouples the two forces at known locations is a redundant problem. We may leverage the prior estimated values for $\mb{f}_\text{rcm}$ and $\mb{f}_\text{ins}$ to resolve this redundancy. We  then propose to use the following estimator:
\begin{align}
	&\hat{\mb{f}}_\text{case2} =\;  \; 
	{\bs{\Gamma}(t)}^\dagger \; \bar{\bs{\xi}}_b + 
	\left(\mb{I} - {\bs{\Gamma}(t)}^\dagger {\bs{\Gamma}(t)}\right) \; \bs{\rho}	\\[12pt]
	&\bs{\rho} = \; \; \nabla_{(\mb{f}_\text{case2})} \left(\tfrac{1}{2}\big\|\mb{f}_\text{case2} - \mb{f}_\text{case2}^{(t_0)}\big\|^2\right) \;\;\Big\rvert_{\mb{f}_\text{case2}=\hat{\mb{f}}_\text{case2}}\\[12pt]
	&\hat{\mb{f}}_\text{case2}\triangleq\;
	\begin{bmatrix}
		\phantom{\Big[}\hspace{-1mm}
		\hat{\mb{f}}_\text{ins} 
		\hspace{-1mm}\phantom{\Big]}\\
		\phantom{\Big[}\hspace{-1mm}
		\hat{\mb{f}}_\text{rcm} 
		\hspace{-1mm}\phantom{\Big]}
	\end{bmatrix}, \quad 
	\bar{\bs{\xi}}_b  \triangleq 
	\begin{bmatrix}
		\phantom{\Big[} \hspace{-2mm}
		-\bar{\mb{f}}_b 
		\hspace{-1mm}\phantom{\Big]}\\
		\phantom{\Big[} \hspace{-2mm}
		-\bar{\mb{m}}_b 
		\hspace{-1mm}\phantom{\Big]}
	\end{bmatrix}
\end{align}
Where $\mb{f}_\text{case2}^{(t_0)}$ can be a trusted estimate in estimator history. The augmented estimated force vector $\hat{\mb{f}}_\text{case2}$ includes both $\hat{\mb{f}}_\text{rcm}$ and $\hat{\mb{f}}_\text{ins}$. The estimator uses measurements of $\bar{\bs{\xi}}_b$ that includes both $\bar{\mb{f}}_b$ and $\bar{\mb{m}}_b$.\\

\section{Simulation Validations} \label{simulations}
\subsection{Simulation Environments}
Following the discussion in the previous sections, we have conducted simulation studies on three different 7-DoF robot models in both the Matlab (two models) and ROS2 (one model) environments to validate the accuracy of the proposed algorithm. Within the Matlab environment, one virtual robot defined by three segments (representing the shoulder, elbow, and wrist) was used. Another simulation in the Matlab environment utilized the Kuka manipulator with its built-in URDF model. A third simulation was conducted in the ROS2 environment to examine the real-time behavior of the manipulator and its custom-designed instrument driving module. 

\begin{figure*}[!h]
	\setlength\Myfigwd{0.73\columnwidth}
	\floatbox[{\capbeside\thisfloatsetup{
			capbesideposition={right,center},
			capbesidewidth=\dimexpr\linewidth-\Myfigwd-2em\relax}}]{figure}[\FBwidth]
	{\caption{Trocar-RCM linear displacement simulation: 1) A surgical room featuring a 7-DoF robot virtual model, representing the shoulder (3 DoF), elbow (1 DoF), and wrist (3 DoF), subject to the patient's body motion, causing a 2cm linear path trajectory of the Trocar-RCM.
			2) The exerted force converges to zero with the proposed admittance controller.}
		\label{virtual_sim}}
	{\includegraphics[width=\Myfigwd]{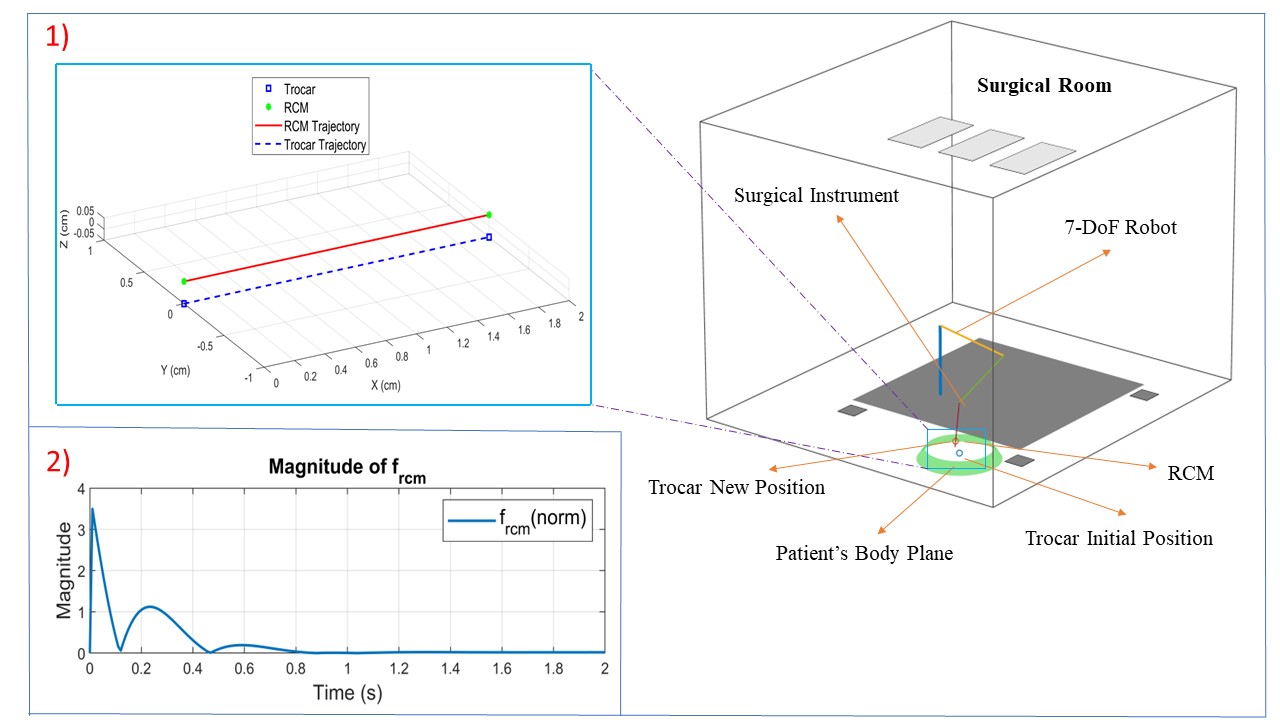}}
	\vspace{-3mm}
\end{figure*}

\subsection{Validating RCM Errors and RCM Force Tracking Performance}

To evaluate the augmented Jacobian method outlined in \eqref{eq:16}, various gain values required in $\mb{G}$ have been tested for the RCM constraint and the instrument command motion. Considering the accuracy of RCM point placement into the trocar, these values have been chosen as follows to stabilize control in \eqref{RCM_velocity2}. 
\begin{align}
	& \mb{K}_\text{ins} = \text{diag}([20, 20, 20]), \\
	& \mb{K}_{\text{adm}\perp} = \text{diag}([0.1, 0.1, 0.1])
\end{align}

The data presented in Fig.\ref{simulation_plots} show that the RCM was aligned with the trocar position and the errors converged. The RCM force tracking was also validated using equations \eqref{eq:16} and \eqref{RCM_velocity2}, as well as the convergence of the estimator $\hat{\gamma}$, to the desired value. This was achieved with the application of external force, as explained in Section\ref{eta variation formula}. \par

\begin{figure}[!b]
	\centering
	\begin{subfigure}[b]{1\columnwidth}
		\centering
		\includegraphics[width=\linewidth]{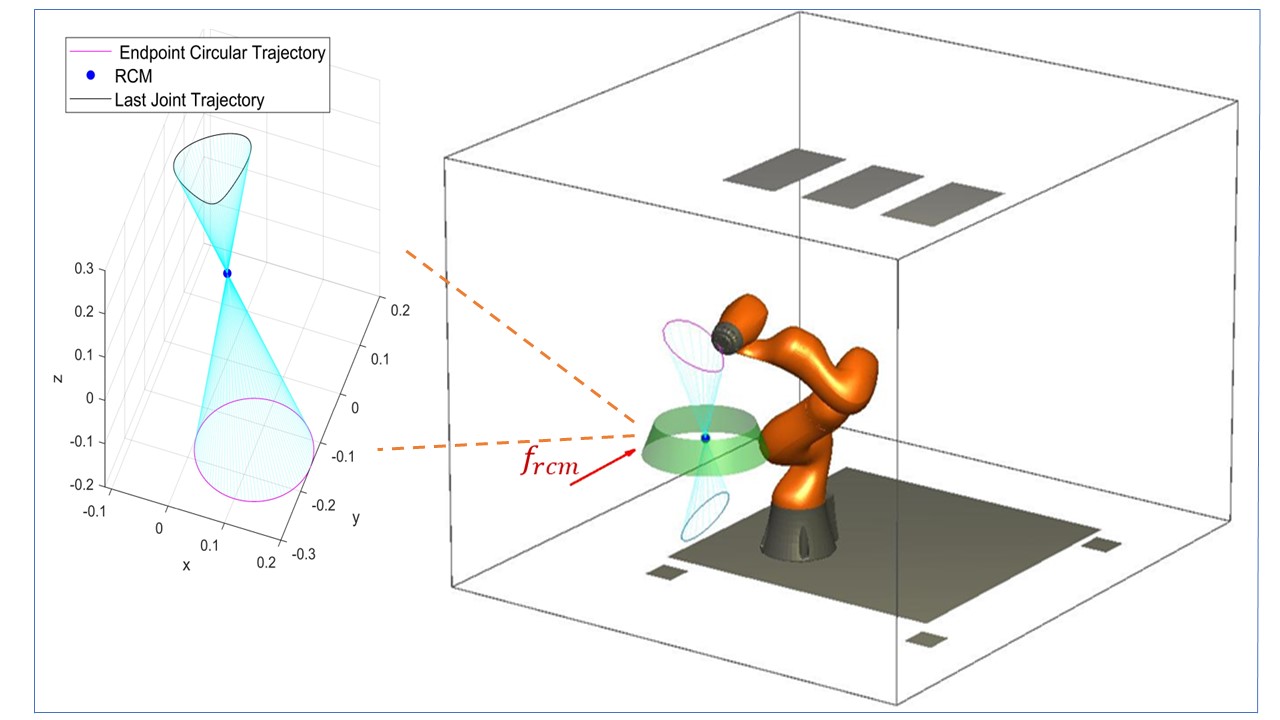}
				\caption{Simulation in MATLAB of the Kuka-iiwa robot's \emph{end-effector} following a circular trajectory, with $\mathbf{f}_\text{rcm}$ being applied at the point.}
		 \label{traject_sim_matlab}
	\end{subfigure}
	\hfill
	\begin{subfigure}[b]{\columnwidth}
		\centering
		\includegraphics[width=\linewidth]{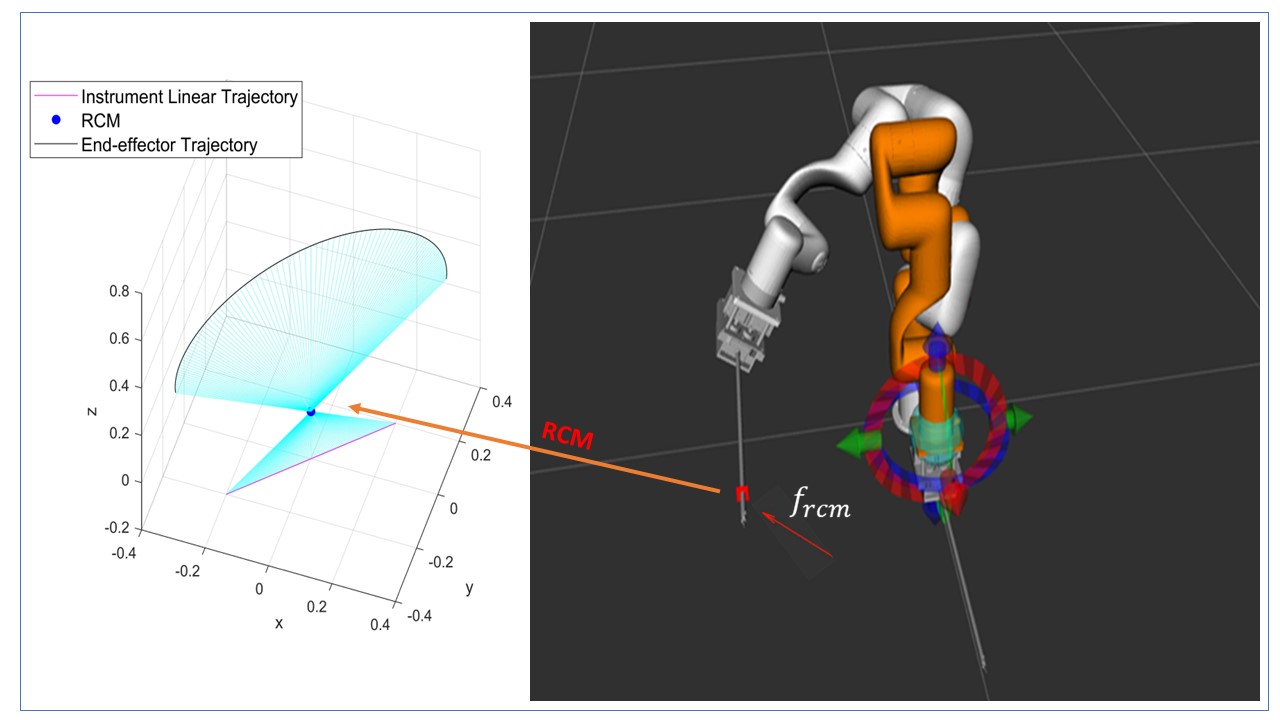}
				\caption{Simulation in ROS2 of the xArm7 robot's \emph{end-effector} following a linear trajectory, with $\mathbf{f}_\text{rcm}$ being applied at the point.}
				\label{traject_sim_ros2}
	\end{subfigure}
	\caption{Simulation results in MATLAB and ROS2: The \emph{instrument} and \emph{end-effector} trajectories are subject to RCM constraints and a continuous exerting force acting upon the instrument.}
	\label{traject_sim}
\end{figure}

In another simulation, the surgical instrument is assumed to be inside the abdominal cavity, as indicated in Fig.~\ref{virtual_sim}-1. We demonstrated that the velocity of the RCM point is proportional to the external force applied. A minor movement of the trocar-RCM along a horizontal line (2cm length) was simulated to replicate the back and forth motion of the abdominal cavity during the repetitive breathing process. The proposed admittance controller, in conjunction with the kinematic framework, was utilized. While the trocar was moving, an external force was applied to the instrument shaft from the patient's side. Subsequently, the force at the RCM location was estimated, using \eqref{force_error}, and then fed into \eqref{eq:16}. The admittance control action was implemented in the inverse kinematics command loop in real-time. The plots reported in Fig.~\ref{virtual_sim} illustrate that the system followed the external forces and accurately kept the RCM point inside the moving trocar positions. \par
An instrument trajectory following simulation was also validated, reported in Fig.~\ref{traject_sim}. A desired circular path was sent for the instrument to follow. Using \eqref{eq:16}, the system was able to follow the circular path while respecting the RCM constraint in presence of external force. As depicted in Fig.~\ref{traject_sim}-a and b, simulations were conducted in both environments using two different 7-DoF robots: Kuka-iiwa and xArm7 in MATLAB, and real-time visualization in ROS2.\par

\section{Future Work}
An experimental hardware system has been prototyped as shown in Fig.~\ref{hw}. The developed robot-assisted MIS system includes a 7-DoF robot manipulator, an Instrument Driving Mechanism, and an off-the-shelf surgical instrument. The software interface has been developed using ROS2, which can leverage the kinematics and control framework presented in earlier sections. In the future, we plan to validate the RCM point tracking with the admittance control on the developed experimental setup. \par
Force estimation at the RCM point is a key component required by the admittance controller. While a mathematical and software framework has been completed, the noise aspects of force sensing has not been considered, which could present challenges when deploying experiments. An observer may be designed and implemented to mitigate this issue. In addition, the current admittance control is a simplified controller with only gains on force feedback. To mitigate the force sensing noise, a more advanced hybrid control law could be designed to leverage the kinematic information such as the estimated RCM position and the estimated trocar position.

\section{Conclusion}
This paper introduces a novel kinematic approach for robot-assisted laparoscopic surgery. The proposed framework can enforce the RCM constraint while mitigating the potential RCM position deviation via an admittance control scheme utilizing external forces at the RCM. 
The admittance controller was integrated into a redundancy resolution augmented Jacobian method. Using force/torque sensory feedback at the base of an instrument driving mechanism (IDM), the proposed framework estimates the forces at RCM in a robust manner that decouples the effect of forces at the instrument distal ends. The estimated RCM force is then used in the admittance controller.  \par
In this paper, we report analysis from kinematic simulations to validate the proposed framework. The implementation of the proposed formulation has yielded promising results in ROS2 and MATLAB. Two approaches have been tested to validate the correctness and accuracy of the method presented in this paper.
In addition, a hardware platform has been completed, and future work is planned for experimental validation.

\clearpage
\bibliographystyle{IEEEtran}
\bibliography{ref}

\end{document}